%% file: root.tex
\documentclass[a4paper, 10pt, conference]{ieeeconf}      

\IEEEoverridecommandlockouts                              

\usepackage{amsmath} 
\usepackage{amssymb}  
\usepackage{hyperref}
\usepackage{cleveref}
\usepackage[per-mode=symbol,detect-all]{siunitx}
\usepackage{booktabs}
\usepackage{todonotes}
\usepackage{pgfplots}
\usepackage[protrusion=true,expansion=true]{microtype} 
\usepackage[noend]{algcompatible}
\usepackage[keeplastbox]{flushend}
\usepgfplotslibrary{groupplots}
\usetikzlibrary{positioning}
\newcommand{\argmax}{\operatornamewithlimits{arg\,max}}

\pgfplotsset{compat=newest}

\usepackage[backend=bibtex,style=ieee, maxcitenames=2, mincitenames=1]{biblatex}
\AtEveryBibitem{
 	\clearfield{url}  
 	\clearfield{doi}  
\ifentrytype{inproceedings}{
 	\clearlist{address}
 	\clearlist{publisher}
 	\clearname{editor}
 	\clearlist{organization}
 	\clearfield{pages}  
 	\clearlist{location}
 	\clearfield{volume}
 }{}
 }

\AtBeginBibliography{\small}

\title{\LARGE \bf
Cooperation-Aware Reinforcement Learning \\
for Merging in Dense Traffic
}

\author{Maxime Bouton,$^1$ Alireza Nakhaei,$^2$ Kikuo Fujimura,$^2$ and Mykel J. Kochenderfer$^1$%
    \thanks{*This work was supported by the Honda Research Institute.}
    \thanks{$^{1}$ Maxime Bouton and Mykel J. Kochenderfer are with the Department of Aeronautics and Astronautics, Stanford University, Stanford CA 94305, USA,
            {\tt \{boutonm,mykel\}@stanford.edu}.}%
    \thanks{$^{2}$ Alireza Nakhaei and Kikuo Fujimura are with the Honda Research Institute, 375 Ravendale Dr., Mountain View, CA 94043, USA, 
            {\tt {anakhaei,kfujimura}@hra.com}.}%
}

\begin{document}

\maketitle
\thispagestyle{empty}
\pagestyle{empty}


\begin{abstract}
Decision making in dense traffic can be challenging for autonomous vehicles. An autonomous system only relying on predefined road priorities and considering other drivers as moving objects will cause the vehicle to freeze and fail the maneuver. Human drivers leverage the cooperation of other drivers to avoid such deadlock situations and convince others to change their behavior. Decision making algorithms must reason about the interaction with other drivers and anticipate a broad range of driver behaviors. In this work, we present a reinforcement learning approach to learn how to interact with drivers with different cooperation levels. We enhanced the performance of traditional reinforcement learning algorithms by maintaining a belief over the level of cooperation of other drivers. We show that our agent successfully learns how to navigate a dense merging scenario with less deadlocks than with online planning methods.
\end{abstract}


\section{Introduction}

Merging into very dense traffic situation is a challenging task for autonomous vehicles. Some decision making algorithms are overly conservative and sometimes fail to create a gap in the traffic. Performing a merge maneuver requires reasoning about the reaction of traffic participants to the merging vehicle in a short time. When no gap is present in the traffic, the autonomous vehicle must be proactive in distinguishing drivers that are willing to slow down and yield to the merging vehicle. 

It has been shown that planning algorithms must consider interaction and joint collision avoidance models to avoid deadlock situations also known as the freezing robot problem~\cite{trautman2010}. Previous research addressed the issue of interaction-aware planning by combining a probabilistic interaction model with an online planner~\cite{ward2017,hubmann2018,schmerling2018, fisac2019}. Online planners simulate the environment up to a certain horizon and take actions that maximize the expected reward of the corresponding simulated trajectories. These approaches can scale to large environments and continuous state spaces, but they still suffer from the curse of dimensionality. The computational complexity associated with dense traffic scenarios limits the planning to short time horizons~\cite{hubmann2018, bansal2018collaborative} or limit the number of vehicles considered~\cite{schmerling2018, fisac2019}. 

The performance of the resulting policies is greatly influenced by the underlying model used to represent the environment~\cite{sunberg2017}. \citeauthor{sunberg2017} showed that a significant improvement in performance can be gained when the planning algorithm has access to information about the driver internal state in lane changing scenarios. Previous work address the problem of modeling interactions between traffic participants using data-driven approaches, probabilistic models, inverse reinforcement learning, rule-based methods, or game theoretic frameworks~\cite{dong2017, sadigh2016,schmerling2018, ward2017,fisac2019}. Inverse reinforcement learning techniques and game theoretic frameworks are generally too computationally expensive to be used in an online planning algorithm considering more than two traffic participants~\cite{sadigh2016, fisac2019}. \citeauthor{schmerling2018} demonstrated a data-driven approach to learn the interaction model on a traffic weaving scenario involving two agents \cite{schmerling2018}. They leveraged parallelization to use this model efficiently for online planning. Such an approach is promising but is not suitable for dense traffic scenarios where more than two traffic participants are interacting. 

Instead of relying on online planning methods, we propose to learn an efficient navigation strategy offline, using reinforcement learning (RL). RL provides a flexibility in the choice of the interaction model. RL has been applied to a variety of driving scenarios such as lane changing \cite{wang2018}, or intersection navigation \cite{tram2018,bouton2019}.

\begin{figure}[!t]
    \centering
    \includegraphics[width=\columnwidth]{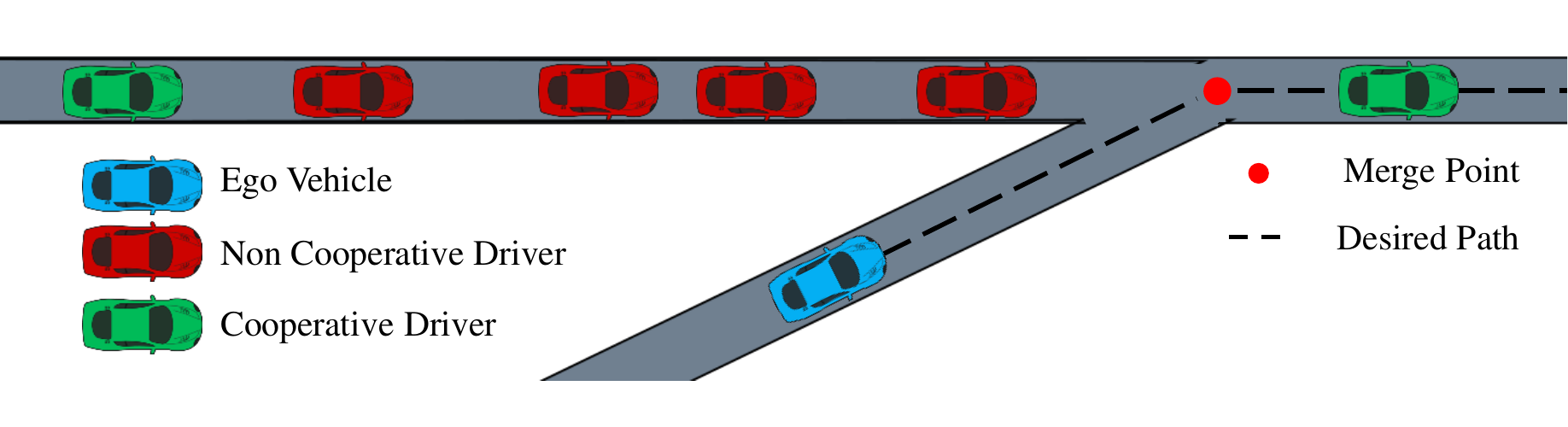}
    \caption{Example of a merging scenario in dense traffic. Drivers on the main road have different reactions to the ego vehicle's motion. Their behavior continuously spans from cooperative drivers that yield to the merging vehicle (green cars) or non cooperative drivers that ignore it (red cars). This paper analyzes how such behavior information can be used in reinforcement learning.}
    \label{fig:scenario}
\end{figure}

In this work, we analyze the ability of an RL agent to benefit from interaction between traffic participants in dense merging scenarios. We show that deep reinforcement learning policies can capture interaction patterns when trained in a variety of different scenarios, even if information about the driver behavior is not available. This approach is then combined with a belief updater that explicitly maintains a probability distribution over the driver cooperation levels. Simulation results shows that an RL agent using belief states as input yields better performance than standard RL techniques as well as an online planning solver. In addition, we propose a simple rule-based behavior parameterized by a cooperation level to model the reaction of vehicles on the main lane to the merging vehicle. 

The scenario of interest is illustrated in \cref{fig:scenario}. Vehicles on the main lane have priority over the merging car (in blue). We focus on dense traffic situation where cars drive slowly (around \SI{5}{\meter\per\second}) and very close to each other (the gaps can be below \SI{2}{\meter}). The ego vehicle on the merging ramp must merge into traffic at a given merge point. When the gaps between vehicles is not sufficient, the merging vehicle can only merge if vehicles on the main lane agree to yield.

\section{Background}

In this section we introduce background material on partially observable Markov decision processes and reinforcement learning. 

\subsection{Partially Observable Markov decision processes}

A partially observable Markov decision process (POMDP) is defined by the tuple $(\mathcal{S}, \mathcal{A}, \mathcal{O}, T, R, O, \gamma)$ where $\mathcal{S}$ is the state space, $\mathcal{A}$ the action space, $\mathcal{O}$ the observation space, $T$ the transition model, $R$ the reward function, $O$ the observation model, and $\gamma$ the discount factor. An agent taking action $a\in\mathcal{A}$ in state $s\in\mathcal{S}$ transitions to a next state $s'$ with probability $T(s, a, s') = \Pr(s' \mid s, a)$ and receives a reward $R(s, a, s')$. In a POMDP, the agent does not know the state, instead, it maintains an internal knowledge of the state through a belief $b$ such that $b(s)$ represents the probability of being in state $s$. The belief is updated at each time step after receiving an observation $o$. The observation is related to the state through the observation model as follows: $O(o, s, a) = \Pr(o \mid s, a)$. 

In a POMDP, actions are chosen according to a policy $\pi$ mapping beliefs to action. Policies are associated to value functions $Q^\pi(b, a)$ representing the expected accumulated reward when taking action $a$ in belief state $b$ and then following policy $\pi$. The agent seeks to maximize the expected accumulated discounted reward. Searching for the optimal value function is often intractable but many algorithms can compute reasonable approximation of the optimal value function~\cite{dmu}.

\subsection{Reinforcement Learning}

In this work, we consider reinforcement learning (RL) as an approximate planning technique to solve POMDPs. RL restricts the search of policies to functions mapping observations to actions instead of mapping beliefs to actions. So-called memoryless policies are often a competitive alternative to belief state policies for solving POMDPs \cite{loch98}. Standard RL makes the underlying assumption that the POMDP is an MDP of state space $\mathcal{O}$. In this MDP, the optimal value function satisfies the Bellman equation:
\begin{equation}
    Q^*(o, a) = R(o, a) + \gamma \sum_{o'}T(o, a, o')\max_a Q^*(o', a)
    \label{eq:bellman}
\end{equation}

In problems with continuous observations, the value function must be approximated. In this work we use a neural network to represent the value function. Such procedure is referred to as deep Q-learning (DQN)~\cite{mnih2015}. The solution to \cref{eq:bellman} can be approximated by the network minimizing the following loss function:
\begin{equation}
	J(\theta) = \mathbb{E}_{o'}[(r + \gamma\max_{a'}Q(o', a'; \theta) - Q(o, a; \theta))^2]
\end{equation}
where $\theta$ represents the parameters of the network.
Given an experience sample $(o, a, r, o')$, the weights are updated as follows:
\begin{equation}
	\theta \leftarrow \theta + \alpha (r+ \gamma\max_{a'}Q(o', a'; \theta) - Q(o, a; \theta))\nabla_\theta Q(o,a;\theta)
\end{equation}
where $\alpha$ is the learning rate, a hyperparameter of the algorithm. In practice, experience samples are stored in a replay buffer after each interaction. The loss function is optimized over mini-batches sampled from the buffer regularly during the training. Mini-batches are sampled using experience replay~\cite{schaul2016}.

\section{Proposed Approach}

This section describes the modeling of the merging scenario as a POMDP. The proposed model is then used as a simulator for training a reinforcement learning agent using Deep Q learning. 

\subsection{Merging Scenario POMDP}

We model the merging scenario as a POMDP with the following definitions of the states, observations, actions, reward and transition model.

\subsubsection{States}
The state of a vehicle corresponds to its physical state as well as its internal state characterizing its behavior. The physical state of a vehicle corresponds to its distance to the merge point, its longitudinal velocity, and its acceleration. In this work, the behavior is characterized by a single parameter $c$, the cooperation level, detailed in \cref{sec:driver-model}. We let $s^i_t = (x^i_t, v^i_t, a^i_t, c^i_t)$ be the state of vehicle $i$ at time $t$. The state of the ego vehicle (the controlled agent), $s^e_t$, does not contain the behavior parameter. The complete state of the environment consists of the collection of the individual states of each vehicle present: $s_t = (s^e_t, s^1_t, \ldots, s^{n_t}_t)$ where $n_t$ is the number of vehicles present at time $t$. Our formulation does not restrain the number of vehicle considered in the problem. 

\subsubsection{Observations}\label{sec:observations}
\begin{figure}
    \centering
    \includegraphics[width=\columnwidth]{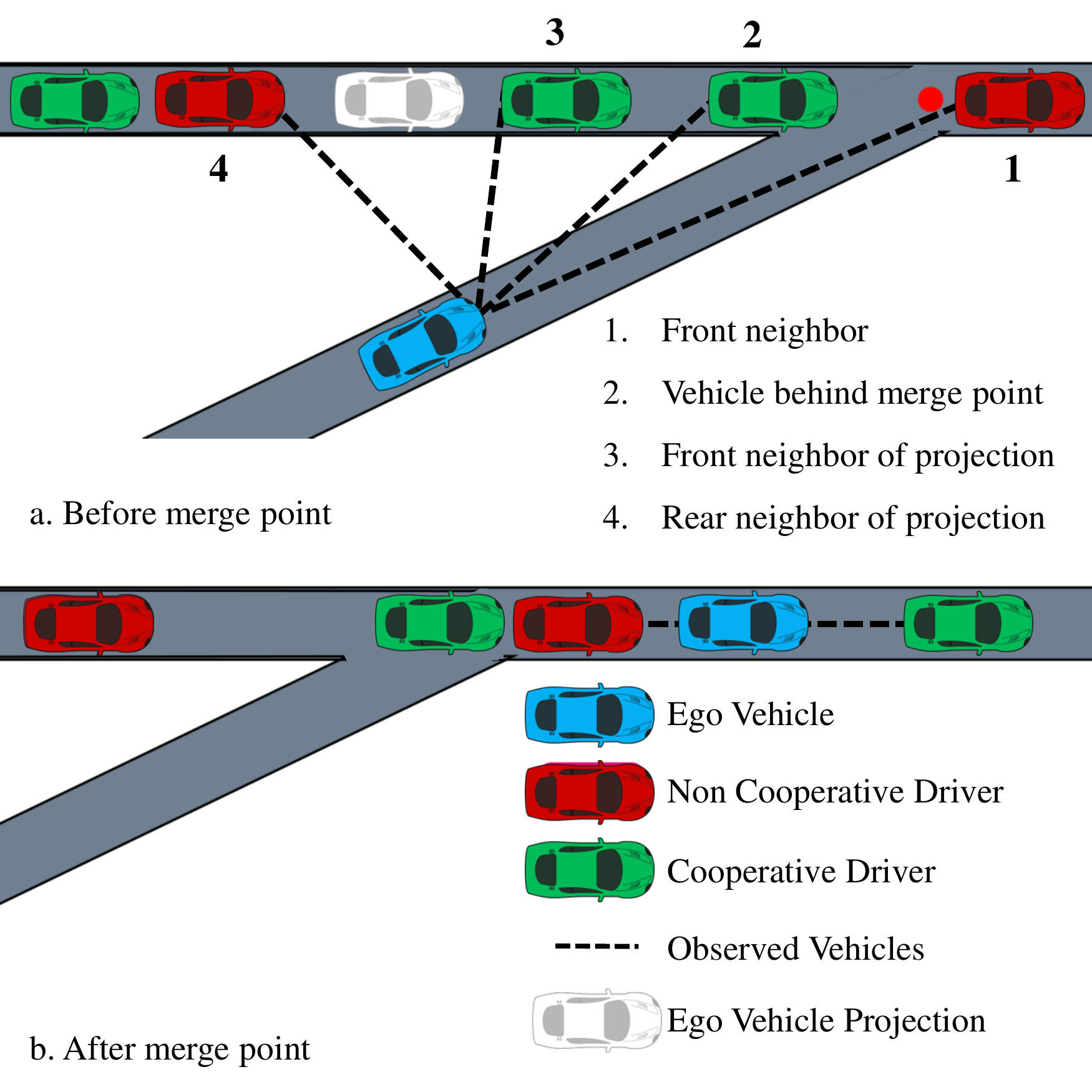}
    \caption{Illustration of the vehicles observed by the ego vehicle. The observation vector (or feature vector) contains information on the position and velocity of the observed vehicles. When less than four vehicles are observed, redundant information is present in the feature vector to preserve its dimension. In addition we analyzed cases where the cooperation level of each observed cars is part of the feature vector.}
    \label{fig:observation}
\end{figure}

The ego vehicle has limited sensing capabilities. It can only sense the vehicles within a certain range, and cannot measure internal states of other vehicles. In this work we do not consider sensor noise. Instead, we focus on the partial observability introduced by the internal state governing the behavior of other drivers. To simplify the observation space, we restrict the observation to the longitudinal position and velocity of four neighbor cars: the front neighbor of the ego vehicle, the vehicle right behind the merge point, the rear neighbor and front neighbor of the projection of the ego vehicle on the main lane. To project the ego vehicle on the main lane, we place it at the same distance to the merge point than its actual position on the merge lane. The ego vehicle can observe the longitudinal position and velocity of these four vehicles perfectly if they are in the field of view. The longitudinal position, speed, and acceleration of the ego car are observed. In addition, we compare the cases where the internal state is directly observable by the ego vehicle or not leading to two or three dimensions per neighbor cars. The total dimension of the observation space is \num{15} when the internal states are observed and \num{12} otherwise. We will refer to those two cases as RL with full observation (FO), that can observe the internal state, and standard RL that only observes positions and velocities. \Cref{fig:observation} illustrates the vehicles observed by the ego car. In addition, it can observe its own state (three dimensions). This observation vector is used as input to our RL agent.

\subsubsection{Actions} The ego vehicle controls its motion by applying a change in acceleration. At each time step, the acceleration is updated as follows:
\begin{equation}
    a_t = a_{t+1} + \Delta a
\end{equation}
where $\Delta a$ is the action chosen by the agent in the set $\{\SI{-1}{\meter\per\square\second}, \SI{-0.5}{\meter\per\square\second}, \SI{0}{\meter\per\square\second}, \SI{0.5}{\meter\per\square\second}, \SI{1}{\meter\per\square\second}\}$. The agent can also apply a hard braking action and releasing action which instantaneously set the acceleration to \SI{-4}{\meter\per\square\second} and \SI{0}{\meter\per\square\second} respectively. Hence, the action space is discrete with \num{7} possible actions at each time step. The design of the action space is inspired by the work of \citeauthor{hu2019IDAS}~\cite{hu2019IDAS}.

\subsubsection{Reward} The reward function is designed such that the optimal policy maximizes safety and efficiency. The agent receives a penalty of \num{-1} for colliding with other traffic participants and receives a bonuses of \num{1} for reaching a goal positions defined \SI{50}{\meter} after the merge point. The time minimization is incentivized by the discount factor, the sooner the goal is reached, the less the bonus is discounted. We used a discount factor of \num{0.95}.

\subsubsection{Transition}

Each vehicle follows a one dimensional point mass dynamics:
\begin{align}
    x_{t+1} &= x_t + v_t \Delta t + \frac{1}{2} a_t \Delta t^2 \\
    v_{t+1} &= v_t + a_t \Delta t 
\end{align}
where $x_t$ is the position of the vehicle at time $t$, $v_t$ its velocity and $a_t$ its acceleration. 
Other drivers in the environment follows an interactive driver model described in the next section. In addition, when a vehicle passes the end of the main lane it is spawned at the beginning of the lane again. This technique allows to maintain a dense traffic in simulation.

\subsection{Driver Model}\label{sec:driver-model}

To model the behavior of drivers on the main lane, we propose an extension of the Intelligent Driver Model (IDM)~\cite{treiber2000}. Our model controls the longitudinal acceleration of the  vehicle on the main lane while taking into account merging vehicles. In addition to the IDM parameters, we introduce a cooperation level $c \in [0,1]$ which is a scalar parameter controlling the reaction to the merging vehicle state. $c = 1$ represents a driver who slows down to yield to the merging vehicle if she predicts that the merging vehicle will arrive ahead of time.  $c = 0$ represents a driver who completely ignores the merging vehicle until it traverses the merge point and follows standard IDM. We will refer to this model as Cooperative Intelligent Driver Model (C-IDM). 

C-IDM relies on estimating the time to reach the merge point ($TTM$) for the car on the main lane ($TTM_{a}$) and the car on the merge lane ($TTM_{b}$) to decide whether the merging vehicles should be considered or not. Once the time to merge for both vehicles is estimated, three cases are considered:
\begin{itemize}
\item If $TTM_{b} < c \times TTM_{a}$, the vehicle on the main lane follows IDM by considering the projection of the merging vehicle on the main lane as its front car. 
\item If $TTM_{b} >= c \times TTM_{a}$, the vehicle on the main lane follows standard IDM.
\item In the absence of a merging vehicle or far from the merging vehicle, the C-IDM driver follows the standard IDM. \end{itemize}

\begin{figure}
    \centering
    \includegraphics[width=0.9\columnwidth]{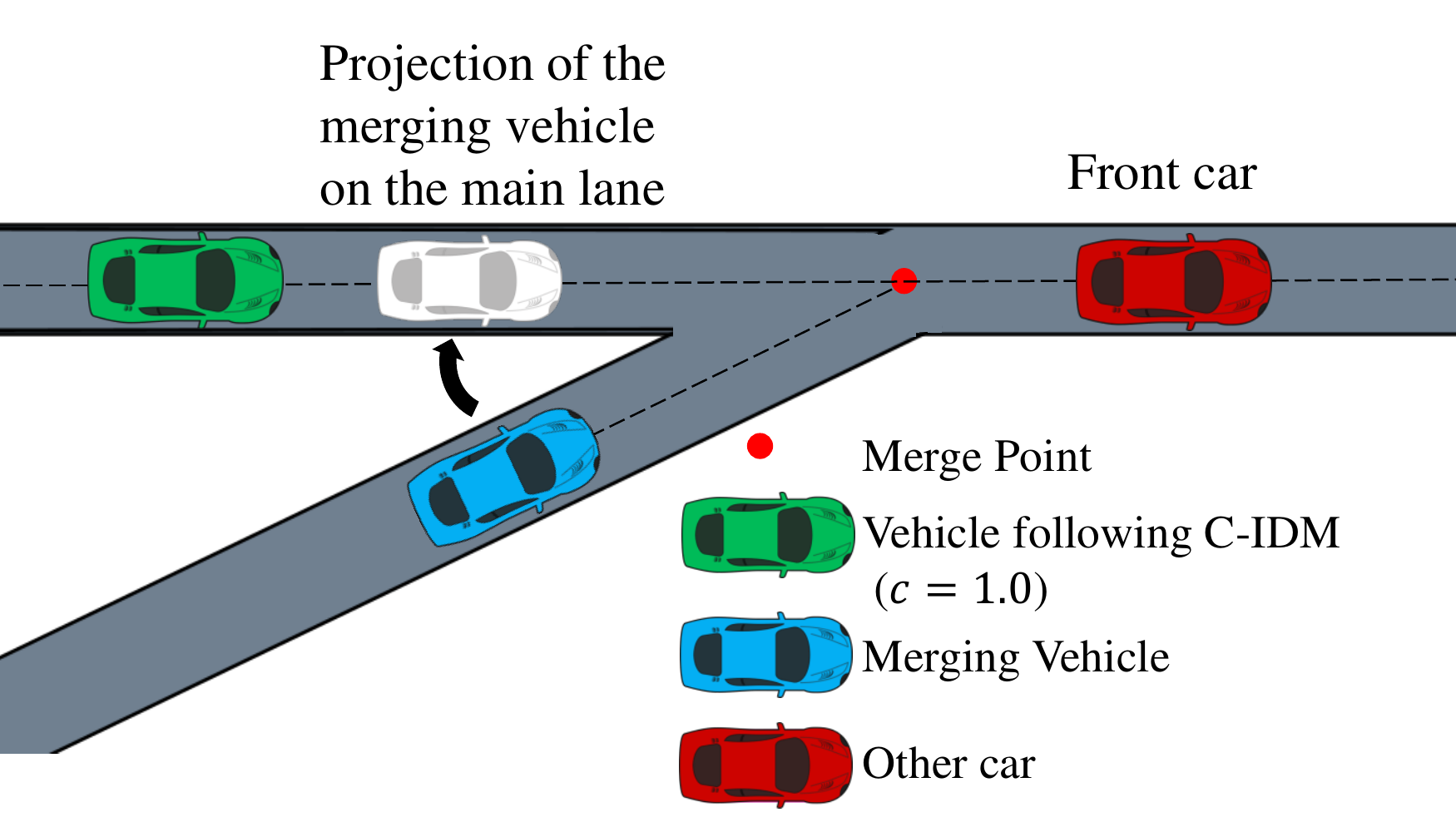}
    \caption{Illustration of the C-IDM model where a cooperative vehicle (in green) considers the merging vehicle as its front car instead of the red vehicle which is the front neighbor used by standard IDM.}
    \label{fig:driver-model}
\end{figure}

Although this model might not represent precisely how human drivers behave, it provides us with a broad range of behaviors by adjusting the cooperation level. In this work we used a simple constant velocity model to estimate the time to merge. A more sophisticated prediction model can be used to have more realistic estimates of the $TTM$. Given the cooperation level, the driver has a deterministic behavior. \Cref{fig:driver-model} illustrates the situation where a cooperative vehicle takes into account the merging vehicle.

\subsection{Inferring the Cooperation Level}

Since the cooperation level cannot be directly measured, the ego car maintains a belief on the cooperation level of the observed drivers. At each time step, the belief is updated using a recursive Bayesian filter given the current observation of the environment.

In this problem, the position and velocity of other drivers is assumed to be fully observable whereas the cooperation level is not always observed. This assumption of mixed observability can help us reduce the computational cost of the belief update. The agent only maintains a distribution over the cooperation level of observed drivers instead of estimating the full state of the environment. Our simple belief updater, is acting as if the cooperation level was binary although it can take a continuous value.  The belief at time $t$ is composed of the fully observable part of the state, $o_t$, and $\theta_i$ for $i=1\ldots n$, where $n$ is the number of observed drivers. $\theta^i_t$ represents the probability that vehicle $i$ has a cooperation level of \num{1}. 
At time $t+1$, the ego vehicle observes $o_{t+1}$ and updates its belief on the cooperation level of vehicle $i$ as follows:
\begin{equation}
    \theta^i_{t+1} = \frac{\Pr(o_{t+1} \mid o_t, c_i=1)\theta^i_t}{\Pr(o_{t+1} \mid o_{t}, c_i=1)\theta^i_t + \Pr(o_{t+1} \mid o_{t}, c_i=0)(1 - \theta^i_t)}
    \label{eq:belief-update}
\end{equation}

\Cref{eq:belief-update} consists of simulating forward the previous state with the two possible hypothesis: $c_i=1$ and $c_i=0$ and comparing the outcome with the current observation. The probability of transitioning from $o_t$ to $o_{t+1}$ given the cooperation level, is computed by propagating the state forward using the proposed transition model and assuming a Gaussian distribution centered around the predicted value with a standard deviation of \SI{1}{\meter} for the positions and \SI{1}{\meter\per\second} for velocities. Without this addition of noise in the transition model, the belief would converge after one observation to $\theta^i = 0$ or $\theta^i = 1$ since the two models would be perfectly distinguishable. 

The focus of this work is not to develop an efficient driver state predictor but rather discuss how this information can be used by an RL agent. Future work might consider more complex filtering techniques such as multi-hypothesis filters, interacting multiple models, or data-driven approaches to estimate the driver cooperation level from observation~\cite{thrun2005}. An additional extension is to consider continuous values of the cooperation level since it is supported by the C-IDM model.

\subsection{Belief State Reinforcement Learning}

Standard reinforcement learning techniques assume that the underlying environment is an MDP. Latent states such as driver behavior characteristics are not explicitly inferred during training. Although memoryless policies can be efficient, reasoning about latent states can often lead to a significant improvement~\cite{sunberg2017}. In the merging scenario, knowing which drivers are more cooperative can help the agent take better decisions. It is expected that the ego vehicle will only try to merge in front of cooperative drivers.

In order to learn such a behavior through reinforcement learning, we propose to use the belief state as input to the reinforcement learning policy. The resulting algorithm is very similar to the standard DQN algorithm applied to the belief MDP. A transition in the belief state MDP can be described as follows:
\begin{itemize}
    \item At time step $t$, the agent has a belief $b_{t-1}$ and receives an observation $o_t$
    \item The new belief $b_t$ is computed using \cref{eq:belief-update}
    \item The agent takes action $a_t = \argmax_a Q(b_t, a)$
\end{itemize}
The rest of the algorithm is identical to standard DQN.

The input to the network is a vector of dimension \num{15}: 
\[
b_t = [o_t, \theta^1_t, \ldots, \theta^n_t ]
\]
where $n$ is the number of observed vehicle, and $o_t$ is the fully observable part of the state (information on position and velocity), and $\theta^i_t$ is the probability of driver $i$ being cooperative. By feeding in the probability on the internal state, the agent can reason in the belief space rather than in the observation space. 
The resulting Q network is combined with a belief updater at test time to result in a policy robust to partial observability.

\section{Implementation}

This section highlights some important aspects of our implementation: the RL training procedure and the distribution of scenarios used during training. Further details can be found in our code base\footnote{\url{https://github.com/sisl/AutonomousMerging.jl}}

\subsection{Reinforcement Learning Training Procedure}

We used a curriculum learning approach to train the agent by gradually increasing the traffic density. When training an RL agent in dense traffic directly, the policy converges to a suboptimal solution which is to stay still in the merge lane and does not leverage the cooperativeness of other drivers. Such a policy avoids collisions but fails at achieving the maneuver. To encourage exploratory actions, we first train an agent in an environment with sparser traffic (\num{5} to \num{12} cars). An alternative to curriculum learning is to design the reward function to incentivize the learning agent to move forward at each time step. However, a more complex reward function often requires a lot of parameter tuning . We found the curriculum learning approach more practical in this work. 

The parameters of the DQN algorithm are summarized in \cref{tab:dqn-params}. It was implemented using the Flux.jl library~\cite{innes2018}. Training one policy took around \num{40} minutes on three million examples.

\begin{table}[h]
	\centering
	\caption{Deep Q-learning parameters}
	\begin{tabular}{@{}ll@{}}
		\toprule[1pt]
		Hyperparameter & Value \\
		\midrule
		Neural network architecture & 2 dense layers of 64 and 32 nodes \\
		Training steps & \SI{3e6}{} \\
		Activation functions & Rectified linear units \\
		Replay buffer size & \SI{400}k experience samples \\
		Target network update frequency & \SI{5}k episodes \\
		Discount factor & \SI{0.95}{} \\
		Optimizer & Adam \\
		Learning rate & \SI{1e-4}{} \\
		Prioritized replay \cite{schaul2016}  & $\alpha=$\SI{0.7}{}, $\beta=$\SI{1e-3}{} \\
		Exploration strategy & $\epsilon$ greedy \\
		Exploration fraction & \SI{0.5}{} \\
		Final $\epsilon$  & \SI{0.01}{}  \\ 
		\bottomrule[1pt]
	\end{tabular}
	\label{tab:dqn-params}
\end{table}

\subsubsection{Initial State Distribution}\label{sec:initialstate}

To populate the initial state of the merging scenarios, we used a two step procedure. 
The first step consists of sampling the number of vehicles present from a desired range. We considered two ranges corresponding to different traffic conditions:
\begin{itemize}
    \item Mixed traffic: between \num{5} and \num{12} cars on the main lane. The agent will experience both sparse traffic scenarios and dense traffic scenarios. 
    \item Dense traffic: between \num{10} and \num{14} cars on the main lane. 
\end{itemize}
The main lane has a length of \SI{150}{\meter} and the vehicles have a length of \SI{4}{\meter}. In dense traffic situations, the gap between vehicles varies from less than \SI{2}{\meter} to larger distances. 
Once the number of cars is decided, vehicles are randomly positioned on the main lane. The initial velocity of each vehicle is drawn from a Gaussian distribution of mean \SI{5}{\meter\per\second} and a standard deviation of \SI{1}{\meter\per\second} Finally, the desired velocity of each vehicle is drawn uniformly from the set: $\{\SI{4}{\meter\per\second}, \SI{5}{\meter\per\second}, \SI{6}{\meter\per\second}\}$ and their cooperation level is drawn uniformly in the interval $[0, 1]$. The desired velocity is used to parameterize the IDM part of C-IDM.

The second step of the initialization consists of simulating the initial state for a burn-in time randomly chosen between \SI{10}{\second} and \SI{20}{\second}. During this burn-in time, the ego vehicle is not present and the vehicles on the main lane follow the C-IDM. The burn-in time allows the initial state to converge to a more realistic situation. This approach is loosely inspired by the work of \citeauthor{wheeler2015}~\cite{wheeler2015}.

Such procedure ensures that the learning agent experiences a variety of situations during training. The design of the training scenarios is critical to the performance of the RL agent and will determine the ability of the policy to generalize. The more variety it experiences during training, the more the policy can generalize.

\section{Experiments}

To evaluate the ability of our method to use the cooperativeness of other drivers, we compared the performance of different policies in the merging scenario with dense traffic (between \num{10} and \num{14} cars on the main lane). Each policy is evaluated in \num{1000} scenarios with initial states sampled as described in \cref{sec:initialstate}. We measured the percentage of scenarios that ended in collisions (collision rate), the average time to reach the goal position located \SI{50}{\meter} after the merge point, as well as the number of scenarios ending in time-out failure. A time-out failure is declared if the ego vehicle has not reached the goal position after \SI{50}{\second}. Such failure cases are representative of the robot freezing problem~\cite{trautman2010}.

We compare the following methods:
\subsubsection{RL without cooperation level information} The first policy consists of using a standard reinforcement learning algorithm which can observe position and velocity of the other vehicle but not their cooperation level. It is referred to as RL.

\subsubsection{RL with cooperation level information} This policy is trained using standard reinforcement learning but has access to the cooperation level at training and test time. It is referred to as RL (FO).

\subsubsection{belief state RL} This policy is trained in the belief MDP. The input of the policy is the prediction given by the belief updater. It is referred to as belief RL.

\subsubsection{MCTS without cooperation level information} This policy uses the same algorithm as the previous policy but does not have access to information about the cooperation level. Instead, it makes an assumption on the driver cooperation level. We evaluated three different assumptions: 
\begin{itemize}
    \item $c=1$: assumes drivers are always cooperative.
    \item $c=0.5$: assumes a cooperation level of $0.5$, drivers on the main lane will react to the merging vehicle only if it will reach the merge point in twice less time. This behavior is a middle ground between non cooperative and cooperative drivers.
    \item $c=0$: assumes drivers are never cooperative. It is equivalent to assuming that the drivers on the main lane follows IDM and are blind to the merging vehicle.
\end{itemize}

\subsubsection{MCTS with cooperation level information} This policy is using Monte Carlo tree search. To handle the continuous state space we used double progressive widening~\cite{couetoux2011}. We used a computation budget matching the decision frequency of \SI{0.5}{\second}. The exact model of the environment is used to produce the tree, and the root node contains all the information about the cooperation levels. It is referred to as MCTS (FO). We refer the reader to \cite{dmu} for background on the MCTS algorithm.

The first three approaches are offline deep reinforcement learning algorithms where as the last two are online planning methods.

\section{Results and Discussion}

\begin{figure}
    \centering
    \includegraphics[width=\columnwidth]{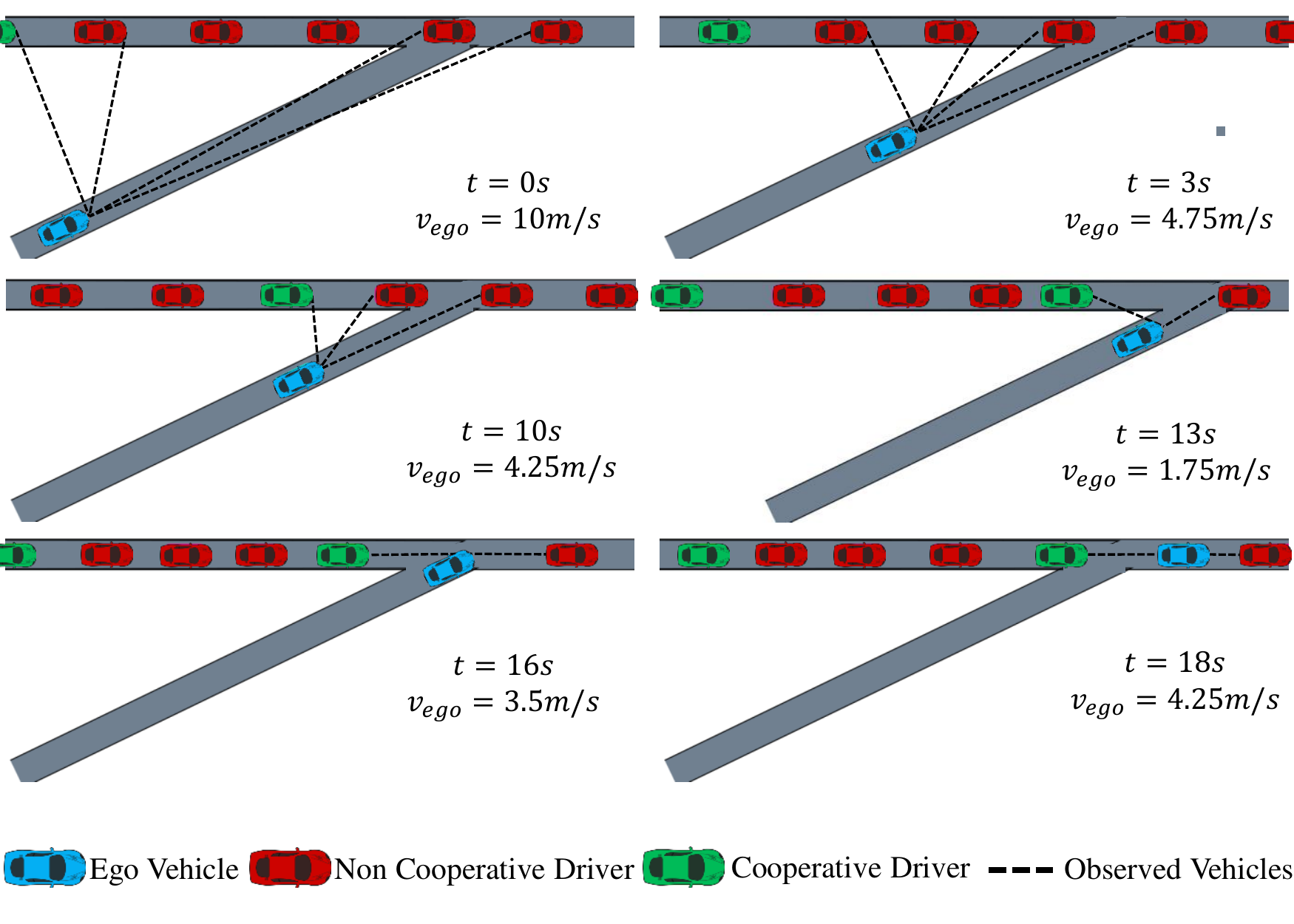}
    \caption{Example of a trajectory when executing the reinforcement learning policy in dense traffic. The ego vehicle learns to merge in front of cooperative drivers.}
    \label{fig:trajectory}
\end{figure}

\begin{figure}
    \centering
    \input{barplots.tex}
    \caption{Performance of the different policies on a dense traffic scenario. Each policy is evaluated on \num{1000} simulations. The policy MCTS (FO) did not result in any collisions. A step in the environment corresponds to \SI{0.5}{\second}. The number of steps corresponds to the average time to reach the goal position.}
    \label{fig:barplots}
\end{figure}
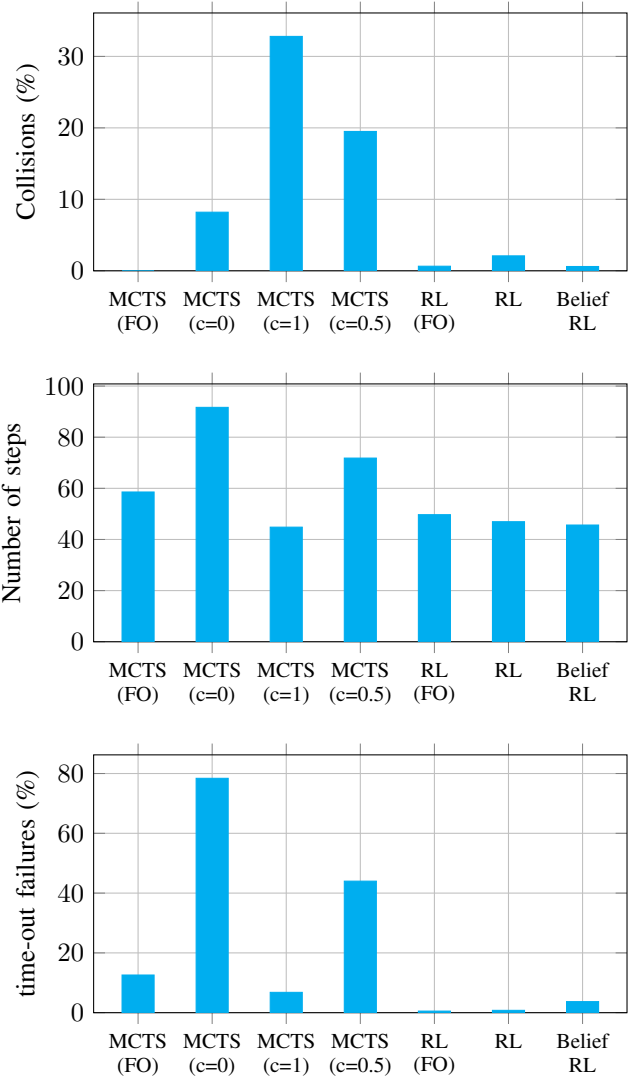

An example of behavior learned through RL is illustrated in \cref{fig:trajectory}. The ego vehicle first slows down as it approaches a dense traffic. Once a cooperative driver is detected ($t=\SI{13}{\second}$) the vehicle slowly merges. At $t=\SI{18}{\second}$ the ego vehicle has merged and follows the traffic on the main lane.

\Cref{fig:barplots} illustrates the performance of the evaluate policy on a dense traffic scenario. We can see from the percentage of collisions that MCTS (FO) is the safest policy, followed by the belief RL approach. The three other MCTS policies had a lot of collisions (much larger than \SI{1}{\percent}). The RL policy without access to information on the cooperation level had \SI{2}{\percent} collisions at test time and the two other RL policies performed similarly with around \SI{0.6}{\percent collisions}. Previous works has shown that only relying on deep RL is not sufficient to achieve safety~\cite{bouton2019}. The deployment of those policies would require the addition of a safety mechanism. It is important to notice that even though they did not have access to the full state, the RL and belief RL policy have much better safety performance than the MCTS approaches that did not have access to this information either.

Regarding the number of time steps to cross, we notice that the MCTS policy with $c=1$ is the most efficient.
The policy is biased towards taking more aggressive actions since it is assuming that every driver is cooperative. On the contrary, the MCTS policy assuming that no driver is cooperative ($c=0$) has a very conservative behavior: it takes the longest time and has the largest number of time-out failures. As expected MCTS ($c=0.5$) has a performance in between the previous two. The RL policies are more efficient than the MCTS (FO) policy and have an average time to cross close to the most aggressive MCTS policy. In addition, we can see that the RL policies have much fewer time-out failures than the MCTS policies. This last fact illustrates that they were able to successfully infer and leverage the information on the cooperation level. 

One can notice that the gap in performance between MCTS (FO) and the other MCTS is much larger than the gap between RL and RL (FO). A possible explanation is that the neural network approximation is able to capture the cooperation level inference task in the hidden layers. However, this implicit state estimation is less efficient than the explicit state estimation provided by combining the belief updater with the RL policy during training and execution. The belief RL policy has a similar safety level and takes, on average, a similar number of steps than the RL policy with perfect observation.

MCTS with full observation still presents a significant number of time-out failures. We believe that relaxing the computation constraint would have lead to better performance. Computation is generally the major bottleneck of online planning algorithms. Our experiments show that the performance can be closely matched using offline trained policies which take a very short time to execute online. However, the MCTS policy with full observation is safer than the deep RL equivalent. A  direction for future work is to use the RL policy as a value estimator to guide the search in MCTS.

\section{Conclusion}

We presented a reinforcement learning approach to the problem of autonomously merging in dense traffic. Our study confirms that an autonomous agent can benefit from reasoning about the interaction with other drivers. We have shown that an agent trained using deep reinforcement learning can outperform online planning methods when being exposed to a broad range of driver behaviors during training. In addition, we presented a belief state RL procedure that explicitly tries to estimate the internal state of other drivers. Finally, we proposed, C-IDM, a simple parametric model capturing a variety of cooperative behaviors. 

Future work involves using more sophisticated techniques to estimate driver behavior. Other algorithms to learn belief state policies could be considered, as well as a direct comparison with online POMDP solvers~\cite{hubmann2018}. Although our RL agent learned more efficient policies, an online planner may provide greater robustness. Using deep reinforcement learning policies to guide the search of a classical planner may be a promising direction.







\printbibliography
\end{document}

%% file: barplots.tex
\begin{tikzpicture}[]
\begin{groupplot}[
    group style={horizontal sep=1cm, 
                 vertical sep=1.5cm, 
                 group size=1 by 3},
    height=5cm,
    width = 8.6cm,
    grid = both ]
\nextgroupplot [
ylabel = {Collisions (\%)}, 
x tick label style={align=center, font=\footnotesize},
ybar=0pt, 
bar width=12pt,
xtick=data, 
symbolic x coords={MCTS \\ (FO), MCTS \\ (c=0), MCTS \\ (c=1), MCTS \\ (c=0.5), RL \\ (FO), RL, Belief \\ RL},
ymin = {0}
]\addplot+ [cyan]coordinates {
(MCTS \\ (FO), 0.0)
(MCTS \\ (c=0), 8.2)
(MCTS \\ (c=1), 32.8)
(MCTS \\ (c=0.5), 19.5)
(RL \\ (FO), 0.64)
(RL, 2.09)
(Belief \\ RL, 0.6)
};
\nextgroupplot [ylabel = {Number of steps},
ybar=0pt,
bar width=12pt,
xtick=data,
x tick label style={align=center, font=\footnotesize},
symbolic x coords={MCTS \\ (FO), MCTS \\ (c=0), MCTS \\ (c=1), MCTS \\ (c=0.5), RL \\ (FO), RL, Belief \\ RL},
ymin = {0}
]\addplot+ [cyan]coordinates {
(MCTS \\ (FO), 58.57)
(MCTS \\ (c=0), 91.673)
(MCTS \\ (c=1), 44.817)
(MCTS \\ (c=0.5), 71.833)
(RL \\ (FO), 49.7)
(RL, 46.95)
(Belief \\ RL, 45.648)
};
\nextgroupplot [ylabel = {time-out failures (\%)}, 
ybar=0pt, 
bar width=12pt,
xtick=data,
x tick label style={align=center, font=\footnotesize},
symbolic x coords={MCTS \\ (FO), MCTS \\ (c=0), MCTS \\ (c=1), MCTS \\ (c=0.5), RL \\ (FO), RL, Belief \\ RL},
ymin = {0}
]\addplot+ [cyan]coordinates {
(MCTS \\ (FO), 12.6)
(MCTS \\ (c=0), 78.4)
(MCTS \\ (c=1), 6.8)
(MCTS \\ (c=0.5), 44.0)
(RL \\ (FO), 0.53)
(RL, 0.78)
(Belief \\ RL, 3.7)
};

\end{groupplot}

\end{tikzpicture}